\documentclass[11pt]{idea_style}
\newif\ifdraft
\draftfalse

\usepackage[T1]{fontenc}
\usepackage{url}
\usepackage{nicefrac}
\usepackage{multicol}
\usepackage{microtype}
\usepackage{geometry}
\usepackage{graphicx}
\usepackage{multirow}
\usepackage{amsmath}
\usepackage{amssymb}
\usepackage{amsfonts}
\usepackage{stmaryrd}
\usepackage{mathrsfs}
\usepackage{xcolor}
\usepackage{textcomp}
\usepackage{manyfoot}
\usepackage{booktabs}
\usepackage{algorithm}
\usepackage{tabularx}
\usepackage{algorithmicx}
\usepackage{algpseudocode}
\usepackage{listings}
\usepackage[inline]{enumitem}
\usepackage{xspace}
\usepackage{makecell}
\usepackage{longtable}
\usepackage{changepage}
\usepackage{color}
\usepackage{colortbl}
\usepackage{pifont}
\usepackage{subcaption}
\usepackage{diagbox}
\usepackage{wrapfig}
\usepackage{siunitx}
\usepackage[misc]{ifsym}
\usepackage{threeparttable}
\usepackage{cutwin}
\usepackage{dsfont}
\usepackage{float}
\usepackage{tablefootnote}
\usepackage{placeins}
\usepackage[numbers]{natbib}

\usepackage{hyperref}
\hypersetup{
  colorlinks,
  linkcolor=ideacolor,
  citecolor=ideacolor,
  urlcolor=ideacolor,
  pdftitle={MIRA: Real-Time Full-Duplex Human-Robot Interaction for Embodied Companions},
  pdfsubject={Real-time embodied companion interaction},
  pdfkeywords={companion robots, embodied interaction, embodiment cues, co-speech motion, streaming motion generation, turn taking}
}
\usepackage[capitalize,nameinlink]{cleveref}

\usepackage{setspace}
\usepackage{changepage}

\ifdraft
    \providecommand\todo[1]{[\textcolor{red}{TODO: {#1}}]}
\else
    \providecommand\todo[1]{}
\fi

\newcommand{\ie}{\emph{i.e.},\xspace}
\newcommand{\eg}{\emph{e.g.},\xspace}
\newcommand{\systemname}{MIRA\xspace}
\newcommand{\rosco}{ROSCO\xspace}
\newcommand{\tagtoken}[1]{\texttt{<motion:#1>}}

\author[1]{Lijian Lin}
\author[1]{Ye Zhu}
\author[1]{Fan Zhang}
\author[1]{Yunfei Liu}
\author[2]{Baofeng Li}
\author[2]{Xianwen Zeng}
\author[2]{Jianan Wang}
\author[1]{Yu Li}

\affiliation[1]{International Digital Economy Academy}
\affiliation[2]{Astribot}

\ideadata[Homepage]{\url{https://gagajian.github.io/MIRA/}}

\title{MIRA: Real-Time Full-Duplex Human-Robot Interaction for Embodied Companions}

\abstract{
\small{
Real-time embodied companion interaction requires a robot to infer user intent from streaming speech, generate timely responses, and execute expressive, interruptible motions. Existing systems typically decouple dialogue orchestration from gesture synthesis, relying on offline motion generation from complete audio. This separation leaves open how a deployed robot can dynamically synchronize response content, prosodic timing, and physical safety under incremental inputs and uncertain turn boundaries. We present MIRA, a unified framework for real-time full-duplex embodied companion interaction. Given streaming user speech, dialogue history, and vocal affect, MIRA predicts both the response text and an explicit embodiment cue that routes the response to the appropriate physical behavior. Discrete social behaviors (\eg listening and greeting) are mapped to validated robot trajectories, while speaking responses are accompanied by streaming, generative co-speech motion.
For co-speech motion generation, we propose ROSCO, a prefix-conditioned diffusion model for streaming audio-to-joint motion generation. We further design RHPC, an inference scheme that maintains a sufficiently long temporal context for motion prediction while bounding physical commitment to a short, interruptible prefix.
At the interaction level, we design CORTEX, a dual-timescale interaction policy that combines low-latency barge-in preemption and streaming response generation with deliberative turn decisions, backed by a robot-side execution layer that enforces physical safety constraints during execution. MIRA is deployed on an Astribot S1 humanoid robot. 
Quantitative evaluations demonstrate competitive audio-motion alignment relative to state-of-the-art motion-generation baselines, while real-robot deployment measurements characterize streaming responsiveness and interruption handling.
}
}

\begin{document}

\maketitle

\begin{figure}[t]
  \centering
  \includegraphics[width=\linewidth]{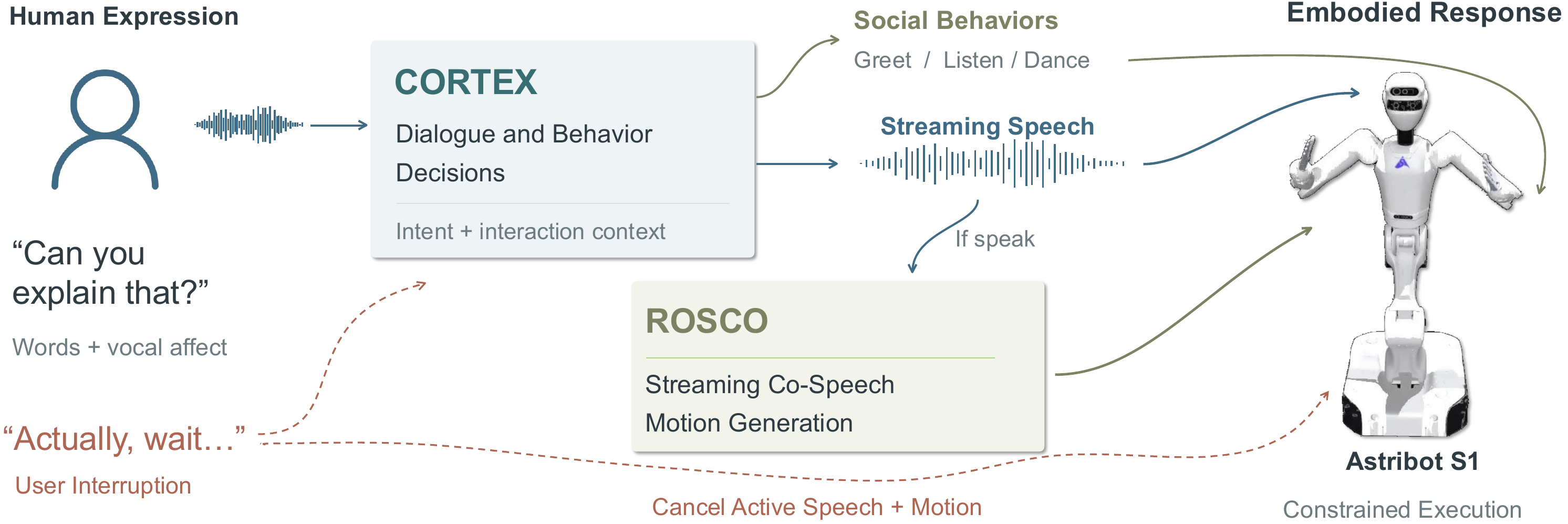}
  \caption{\textbf{Real-time embodied companion interaction in MIRA.}
  Given user speech containing linguistic content and vocal affect, CORTEX infers conversational intent and context.
  Responses are routed into two parallel pathways: discrete social behaviors (\eg greeting, listening) execute pre-validated action libraries, while open-ended speaking drives \textbf{ROSCO} to generate streaming, speech-synchronized motion. When a user barge-in occurs (red dashed path), active speech and joint trajectories are immediately preempted and cancelled at the execution boundary, enabling safe physical halting while retaining conversational context for subsequent turn arbitration.}
  \label{fig:teaser}
\end{figure}

\section{Introduction}
\label{sec:intro}

A companion robot is not only expected to answer a user's questions, but also to participate in an ongoing interaction through speech, listening, and physical behavior. A greeting may call for a brief, recognizable action, and an explanation may call for gestures that follow the rhythm and duration of unfolding speech. As the conversation evolves, the robot must continuously decide which behavior is appropriate, when to speak, how to accompany its speech with motion, and how to respond when the user pauses, asks for clarification, or interrupts. These requirements make companion interaction a continuous embodied process rather than a sequence of isolated speech responses. On a physical robot, this process is further constrained by the fact that motion already released for execution cannot be changed as freely as uncommitted text or audio.

This coordination becomes particularly challenging when speaking behavior must be generated online. Unlike offline co-speech synthesis, the complete response audio may not yet be available when the robot needs to begin moving. The robot must therefore generate motion from incremental audio while maintaining continuity with its own previously generated motion. Meanwhile, the conversational state may change before the response is complete: a user may interrupt, request clarification, or redirect the interaction. A practical companion system must consequently handle two coupled processes: deciding \emph{what the robot should do next} at the interaction level, and continuously generating \emph{how that behavior should be expressed} at the motion level.
The generated trajectories must additionally satisfy robot-specific kinematic and execution constraints.

Prior work has established the importance of coordinating speech, gesture, and conversational state in embodied interaction~\cite{cassell2000eca,skantze2021turntaking}. Co-speech motion generation has subsequently achieved increasingly expressive and realistic behavior~\cite{ginosar2019gesture,ao2022rhythmic,liu2022beat,yi2023talkshow,liu2024emage}, while recent systems have begun to consider streaming behavior generation and robot embodiment. ProAct~\cite{zhang2026proact} combines streaming behavior generation with deliberative social reasoning, PhysDrift~\cite{liang2026physdrift} studies robot-native generation under embodiment constraints, and RoboGesture~\cite{wang2026robogesture} generates streaming, semantically aligned co-speech gestures on a physical humanoid. 
Building on these advances, we explore how these capabilities can be coordinated within full-duplex embodied interaction, where conversational states evolve dynamically and physical behavior must respond continuously to user input. Specifically, we focus on three core challenges: (1) robust conversational turn arbitration that determines whether the user is speaking to the robot and handles interruptions accordingly; (2) low-latency physical preemption that promptly halts ongoing hardware execution upon user barge-in; and (3) seamless integration of discrete social behaviors and continuous, streaming co-speech motion generation.

To address these challenges, we propose MIRA (Real-Time Full-Duplex Human–Robot Interaction for Embodied Companions), which integrates dialogue control, behavior routing, and robotic streaming co-speech generation. MIRA treats physical behavior as part of the interaction policy rather than as a downstream rendering step. 
To connect dialogue decisions with physical behavior, MIRA introduces an \emph{embodiment cue}, a compact symbolic label that specifies how an admitted response should be physically expressed.
The cue routes the response to either a set of validated social behaviors, such as greeting and listening, or the generative speaking pathway. This explicit routing allows social behaviors with well-defined physical requirements to remain reliable while open-ended speech can be accompanied by generative motion. 
The interaction controller, CORTEX, combines streaming response generation with deliberative turn admission and a local interruption gate, allowing the robot to continuously arbitrate when to speak, listen, or yield a turn. Fig.~\ref{fig:teaser} presents the full MIRA framework.

For open-ended speaking behavior, we present \textbf{ROSCO (Robotic Streaming Co-speech Generator)}, which generates robot joint trajectories from streaming audio and recent motion history. ROSCO is designed for recurrent streaming generation, where motion must remain temporally coherent while adapting to newly arriving speech. 
To connect generation with physical execution, we introduce Receding-Horizon Prefix Commitment (RHPC), which separates the prediction horizon from the physical emission stride: the model looks ahead over a longer motion window while releasing only a short prefix for execution. This bounded commitment preserves a longer temporal context for generation while keeping the physically committed motion short enough to remain responsive to interruptions and changes in conversational state.

We deploy MIRA on an Astribot S1 humanoid robot and evaluate ROSCO on retargeted co-speech motion dataset BEAT~\cite{liu2022beat}. Our evaluation combines motion-quality comparisons, streaming timing measurements, and interruption analysis. ROSCO achieves competitive speech-motion alignment while meeting the streaming emission budget under the evaluated deployment configuration. These results, together with qualitative real-robot demonstrations of streaming co-speech motion and barge-in handling, demonstrate the feasibility of MIRA for real-time embodied companion interaction.

The contributions of this work are:
\begin{itemize}
   \item We propose \textbf{MIRA}, a real-time full-duplex human-robot interaction framework that coordinates dialogue decisions, social behaviors, and streaming co-speech motion through an explicit embodiment-cue protocol and unified response-lifecycle management.

    \item We introduce \textbf{CORTEX}, a comprehensive interaction architecture bridging high-level cognition to low-level motor control: it actively generates streaming verbal responses and embodiment cues, manages dual-timescale turn arbitration, and is grounded by a robot-side execution layer enforcing high-rate collision checks, joint constraints, and safe physical halting.

    \item We propose \textbf{ROSCO}, a prefix-conditioned diffusion model for streaming audio-to-joint motion generation, together with \textbf{RHPC}, an inference scheme that balances a sufficiently long temporal context for motion prediction with bounded, interruptible physical commitment.

    \item We present a \textbf{system-level evaluation} of motion quality, streaming performance, and interruption handling, demonstrating these capabilities on a physical Astribot S1 humanoid robot.
\end{itemize}

\section{Related Work}
\label{sec:related}

\subsection{Streaming and Full-Duplex Spoken Dialogue Systems}
\label{subsec:streaming_models}

Recent advancements in spoken dialogue systems have shifted from offline, cascade-based perception toward reactive, streaming interaction. The Audio Interaction Model (AIM)~\cite{xie2026audiointeraction} formalizes this shift as an always-on perceive-decide-respond loop, where a model continuously updates context, decides whether intervention is warranted, and responds without stopping perception. This direction builds on longstanding work on conversational agents and human--robot turn-taking, including endpointing, overlap management, and incremental response timing~\cite{cassell2000eca,skantze2021turntaking}. Related multimodal conversational-agent research also treats speech, visual context, and social behavior as jointly evolving signals rather than isolated pipeline stages~\cite{breazeal2003emotion,paiva2017empathy}. This framing is closely related to our setting: a companion robot cannot wait for a complete, fixed input before choosing whether to speak, hold, or stop.
MIRA extends this streaming paradigm from purely auditory response timing to physical embodiment, where every verbal response is paired with an explicit embodiment decision, discrete social behaviors are routed to validated motion libraries, and open-ended speaking is accompanied by streaming co-speech motion under a bounded, interruptible commitment.

Embodied social agents face an inherent temporal tension between high-level cognitive reasoning and low-latency motor control. While dual-system frameworks such as ProAct~\cite{zhang2026proact} decouple behavioral streaming from cognitive planning to inject proactive intentions, they primarily focus on one-way intention steering and leave open the physical commitment problem during sudden conversational barge-ins. In contrast, MIRA targets full-duplex, bi-directional interaction. Beyond proactive turn decisions, MIRA explicitly bounds physical commitment via RHPC and a robot-side execution bridge to enable safe, low-latency preemption, while bridging dialogue arbitration and execution through an inspectable, symbolic Embodiment Cue interface rather than an unconstrained end-to-end trajectory generator.

\subsection{Embodied Companion and Affective Interaction}
\label{subsec:embodied_companion}

Embodied conversational agents have established that speech, gaze, gesture, facial expression, and dialogue state should be planned as components of a unified interactive behavior rather than as decoupled outputs~\cite{cassell2000eca,cassell2001beat,breazeal2003emotion}. Research on social robot gesture generation likewise shows that communicative motion influences both the interpretation and perceived naturalness of a robot's response~\cite{salem2012gesture,yoon2019robots}. Turn-taking research further demonstrates that endpointing, overlap, and barge-in are central to conversational fluency in human--robot interaction (HRI)~\cite{skantze2021turntaking}. Companion robots, however, introduce a unique physical commitment problem: delayed gestures, stale motion packets, and incorrectly cancelled responses remain physically visible and socially disruptive even after the high-level dialogue state has transitioned.

Affective computing treats vocal, facial, and physiological observations as uncertain evidence about a user's internal state rather than direct ground truth~\cite{picard1997affective}. Social robotics likewise emphasizes that emotional expression must remain legible and appropriate to the specific interaction context~\cite{breazeal2003emotion,paiva2017empathy}. This concern is also reflected in work on expressive face and body representations, which models identity, pose, and affect as structured but uncertain components of animation~\cite{li2017learning,danve2022emoca,pavlakos2019expressive,yin2025smplest}. MIRA aligns with this conservative paradigm. Instead of mapping noisy vocal affect directly to low-level motor commands, the speech-derived affect cue is fused with transcript semantics, dialogue history, and interaction state to select an appropriate high-level embodiment cue, preserving safety and inspectability.

\subsection{Language-to-Behavior Interfaces}
\label{subsec:lang_to_behavior}

Language-conditioned robotics commonly decouples high-level reasoning from low-level joint execution. SayCan combines language-model scores with learned affordances to select discrete robot skills from a pre-defined library~\cite{ahn2022saycan}, while PaLM-E incorporates embodied observations directly into a multimodal foundation model for task planning~\cite{driess2023palme}. This line of work is complemented by approaches that learn cross-modal associations between language and gesture, ground language in embodied observations, and organize robot behavior through reusable skills~\cite{liu2022learning,ahn2022saycan,driess2023palme}. MIRA applies this decoupling philosophy specifically to social response behavior. The predicted embodiment cue describes how the robot should physically accompany the generated response at a semantic level, rather than specifying the low-level joint trajectory itself. This compact interface keeps the dialogue decision highly inspectable and allows downstream motion backends to be modified or upgraded without altering the high-level CORTEX interaction policy.

\subsection{Co-Speech Motion for Avatars and Robots}
\label{subsec:co_speech}

Rule-based and template-driven systems such as BEAT map linguistic structure directly to communicative gesture specifications~\cite{cassell2001beat}. Conversely, data-driven co-speech methods learn high-dimensional motion distributions from multi-modal inputs including audio, text, speaker identity, and prosodic features~\cite{ginosar2019gesture,yoon2020trimodal,ao2022rhythmic}. Semantic and affective annotations further expose the complex one-to-many relationship between an utterance and its possible gestures~\cite{zhang2024semanticgesticulator,liu2022beat}. Dataset and representation studies have also expanded the scope of speech-driven behavior modeling from speaking styles and facial motion to expressive whole-body motion~\cite{cudeiro2019capture,wang2020mead,loper2015smpl,pavlakos2019expressive,lin2023one}. Related reconstruction methods provide structured estimates of body pose, shape, and expressive components that are useful for building and evaluating motion representations~\cite{kolotouros2019learning,kocabas2021pare,Kolotouros2021ProHMR}. Recent generative models synthesize holistic face, hand, and body motion for virtual characters~\cite{yi2023talkshow,liu2024emage,liu2025gesturelsm,mughal2026miburi}, while speech-driven facial animation and head-pose methods explore diffusion-based generation for expressive facial movement~\cite{sun2024diffposetalk}.

Physical robot deployment raises additional kinematic and safety constraints beyond virtual avatar synthesis. Humanoid robot gesture generation highlights the value of physical embodiment and spatial presence~\cite{yoon2019robots,salem2012gesture}, while human-body representations and general motion-retargeting methods provide important foundations for transferring expressive motion across embodiments~\cite{loper2015smpl,pavlakos2019expressive,araujo2025retargeting}. PhysDrift identifies an embodiment gap when human-centric, SMPL-X-style motion pipelines are retargeted to robots, arguing for robot-native co-speech generation under strict joint constraints~\cite{liang2026physdrift}. RoboGesture further learns a streaming diffusion policy in humanoid joint space, aligning multi-granular audio cues with co-speech motion and applying an MPC filter for collision-aware execution on a physical humanoid~\cite{wang2026robogesture}. MIRA shares this robot-native streaming goal, but treats co-speech generation as one pathway inside a full-duplex interaction policy: CORTEX arbitrates turns and barge-in, embodiment cues route between validated discrete social behaviors and generative speaking motion, and RHPC bounds how much predicted motion may be physically committed.

\subsection{Streaming Diffusion and Physical Execution}
\label{subsec:streaming_diffusion}

Diffusion models capture the highly multimodal distribution of plausible communicative gestures~\cite{ho2020ddpm,zhu2023diffgesture,chen2024diffsheg}, and continuous rotation representations help maintain stable orientation trajectories in neural motion synthesis~\cite{zhou2019rotation}. Diffusion Transformers provide a flexible denoising backbone for streaming deployment~\cite{peebles2023dit}, while recent work explores faster latent or masked generation for reducing the cost of iterative synthesis~\cite{liu2025gesturelsm,yang2025gesturehydra,liu2024emage}. However, strong offline clip synthesis does not define how much motion a robot may safely commit before the current response and turn boundaries are fully resolved. Streaming physical deployment introduces causal audio input, generated motion history, cold-start constraints, strict emission deadlines, interruption recovery, and robot-side safety checks.

MIRA treats these challenges as core components of the motion interface rather than mere implementation details. Under RHPC, \rosco predicts an overlapping future horizon from causal audio and generated motion history, but commits only a short prefix for execution. Retrieved discrete motions and generated continuous trajectories then converge at a robot-side execution boundary that validates session identity, checks physical constraints, converts timing, and dispatches commands. 

\section{MIRA System Architecture and Interaction Policy}
\label{sec:system}

\begin{figure}[t]
\centering
\includegraphics[width=\linewidth]{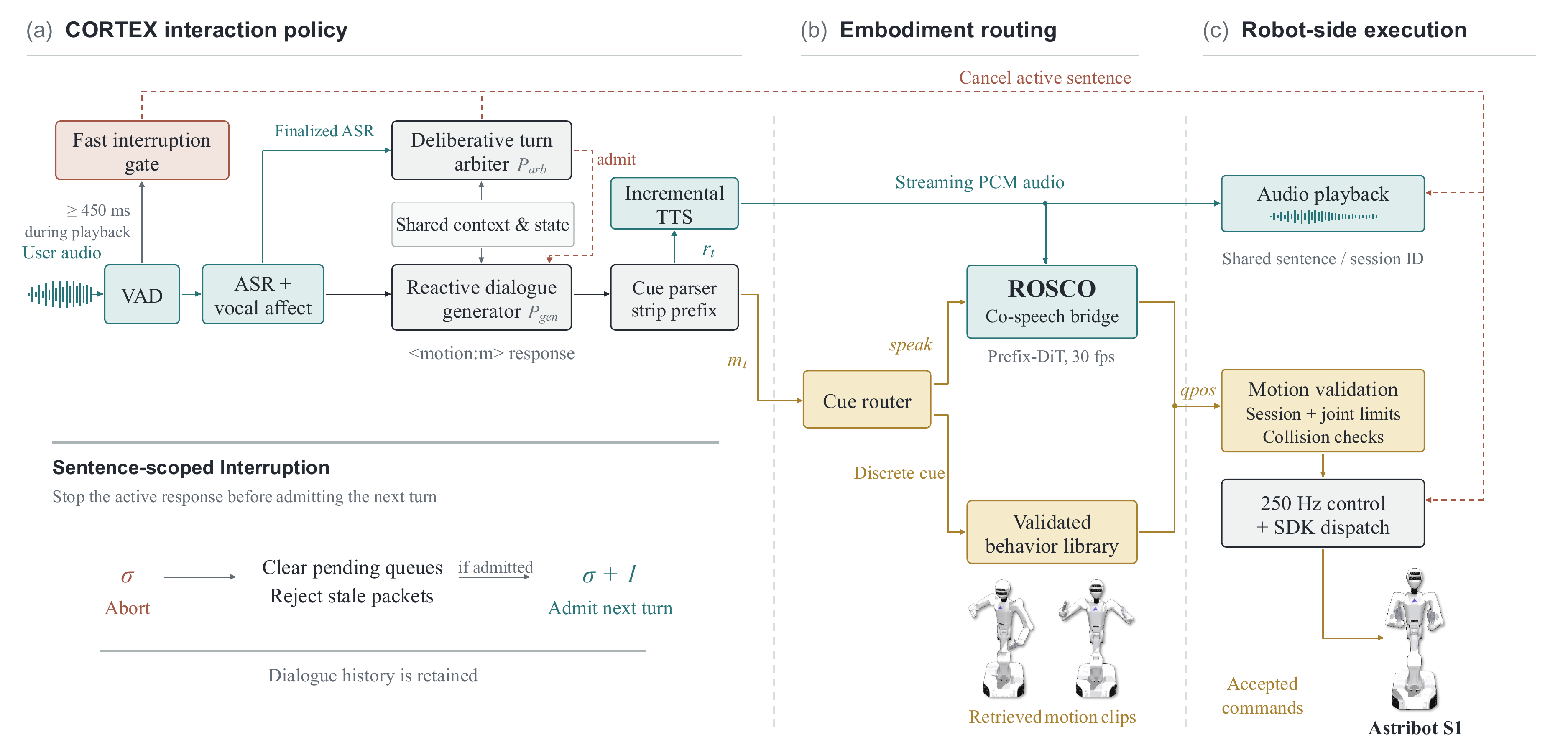}
\caption{\textbf{MIRA system architecture.}
(a) CORTEX combines the transcript, vocal affect, dialogue history,
and playback state to manage turn admission and stream responses.
The cue parser separates embodiment metadata from response text
before TTS.
(b) Embodiment cues route motion production to ROSCO for
audio-conditioned co-speech generation or to a validated behavior
library for discrete actions.
(c) The robot-side execution layer applies configured joint and
collision checks before SDK dispatch and manages audio playback.
Speech and motion share a response identity; the red dashed paths
indicate cancellation requests for the active response, allowing
its pending outputs to be invalidated while retaining dialogue history.}
\label{fig:system_overview}
\end{figure}

\subsection{System Overview}
\label{subsec:MIRA_overview}

MIRA provides a real-time orchestration layer that coordinates user intent understanding, response generation, and embodiment-cue selection with physical execution, with ROSCO (Sec.~\ref{sec:generation}) serving as the streaming co-speech motion generator for open-ended speaking behavior.
To ensure fluid and robust interaction, the system is designed around four requirements. First, interaction decisions must be \emph{contextually appropriate}: MIRA jointly considers the user's current intent, vocal affect, dialogue history, and interaction state to determine whether and how to respond, including whether to interrupt ongoing behavior and which embodied behavior to select.
Second, cue-to-motion routing must be \emph{inspectable}: the dialogue model selects a behavior class rather than directly controlling robot joints. Third, response generation and embodiment must be \emph{streaming}: speech and motion should begin before the complete response is available. Fourth, physical execution must be \emph{contained}: generated and retrieved motion are validated by the robot-side bridge for safe execution.

As illustrated in Figure~\ref{fig:system_overview}, MIRA is organized around three primary functional stages: (a)~\textbf{CORTEX Interaction Policy}, which
makes turn-level interaction decisions and produces the response together
with an embodiment cue; (b)~\textbf{Embodiment Routing}, which routes the selected cue to either a validated behavior library or the ROSCO co-speech pathway; and (c)~\textbf{Robot-Side Execution}, which processes candidate motion under hardware-level constraints and dispatches the resulting motion for execution.
The \textbf{CORTEX Interaction Policy} processes incoming user audio
to obtain linguistic and vocal-affect information, maintains
conversational and interaction state, and produces a natural-language
response together with a symbolic embodiment cue. The internal decision process, including the dual-timescale policies and interruption handling, is described in
Sec.~\ref{subsec:cortex_design}. 
The \textbf{Embodiment Routing} stage maps the selected cue to its
corresponding physical pathway. Discrete social behaviors (e.g.,
\texttt{greeting}, \texttt{listen}) are retrieved from a validated behavior
library, while the \texttt{speak} cue opens a streaming co-speech session
and routes PCM audio to ROSCO. The resulting motion, together with
retrieved behaviors, is passed to the \textbf{Robot-Side Execution} layer,
where session-scoped validation, joint-limit and collision checks,
interpolation, and hardware dispatch are performed. 

\subsection{CORTEX: Dual-Timescale Interaction Policy}
\label{subsec:cortex_design}

CORTEX orchestrates full-duplex embodied dialogue through two cooperative policies for turn admission and response generation. The \textbf{Deliberative Turn Arbiter} ($\mathcal{P}_{\mathrm{arb}}$) determines whether an incoming utterance warrants a response and whether ongoing output should be interrupted, while the \textbf{Reactive Dialogue Generator} ($\mathcal{P}_{\mathrm{gen}}$) produces spoken responses and embodiment cues for admitted turns.
Both policies operate through a shared connection state, ensuring that downstream speech and motion remain associated with a single active response even when decisions arrive asynchronously. Moreover, to reduce latency during active playback, we adopt a low-level, VAD-based \textbf{Fast Interruption Gate} that detects potential user barge-ins before deliberative turn arbitration is completed.

\noindent\textbf{Fast interruption gate.}
To minimize barge-in latency, MIRA uses a local VAD-based\footnote{https://github.com/snakers4/silero-vad} gate to detect sustained user speech during robot playback. After VAD stabilization, the controller measures the duration of the ongoing speech segment. If the segment reaches the configured $450$~ms interruption-confirmation threshold while playback is active and embodiment listening mode is disabled, the controller performs an early abort: it marks the current response as aborted, clears pending TTS output, sends a TTS-stop control message, and terminates the ongoing robot motion. Meanwhile, audio is sent incrementally to the ASR backend while the user is speaking. Deliberative Turn Arbiter is triggered only after the backend returns a provider-specific final or otherwise definitive transcription. If the turn is ignored after an early abort, the controller may generate a continuation of the interrupted response using its saved response context.

\noindent\textbf{Deliberative Turn Arbiter
($\mathcal{P}_{\mathrm{arb}}$).}
After an ASR segment is finalized, the Deliberative Turn Arbiter $\mathcal{P}_{\mathrm{arb}}$ first applies deterministic rules to handle unambiguous cases, including empty or hesitation-only input (\eg 'um', 'hmm') and explicit stop or exit commands. For inputs that require semantic arbitration, it invokes a non-streaming language-model query. The query combines the current transcript, vocal affect, recent dialogue history, and interaction context, including playback status, whether the dialogue remains active, the robot’s current or most recent speech, and the preliminary rule decision. These inputs support context-sensitive judgments about whether an utterance addresses the robot and whether it warrants interrupting the ongoing response. The Arbiter returns one of three semantic decisions:
\begin{equation}
\mathcal{P}_{\mathrm{arb}}
\in \{\texttt{IGNORE},\texttt{REPLY},
\texttt{INTERRUPT\_AND\_REPLY}\}.
\end{equation}
Here, \texttt{IGNORE} declines to admit the input as a new user turn; \texttt{REPLY} admits a turn for response generation; and \texttt{INTERRUPT\_AND\_REPLY} requests interruption of the ongoing response before generating a new one. The controller combines this decision with deterministic rules to determine the final action.

\noindent\textbf{Reactive Dialogue Generator
($\mathcal{P}_{\mathrm{gen}}$).}
Once a turn is admitted, CORTEX invokes the Reactive Dialogue Generator ($\mathcal{P}_{\mathrm{gen}}$) to interpret the user's request and formulate a response.
Generation is conditioned on the current transcript, vocal affect, accumulated dialogue history, and configurable persona and task instructions. The persona and task instructions specify the assistant's identity, conversational style, and interaction role, supporting companion dialogue and informational presentations. Based on the inferred intent, the generator can invoke registered functions, such as weather lookup and music playback. Tool calls and their results are recorded in the dialogue context, and returned results inform subsequent response generation when needed. 
In addition, the model is prompted to emit a symbolic embodiment cue, which is parsed separately from the spoken text to select the corresponding embodiment route. The response text is streamed to incremental TTS,
allowing speech and associated co-speech motion to begin before the complete response is available.

\noindent\textbf{State handoff and interruption semantics.}
CORTEX coordinates turn arbitration and response generation
through shared connection state, including playback status,
an abort flag, and the current response identifier $\sigma$.
When a turn is admitted, the generator creates a response with a fresh identifier, which associates queued TTS segments and motion-control messages with that response.
An interruption marks the current response as aborted, clears pending output queues, and dispatches speech-stop and motion-end commands. These operations stop ongoing response production while retaining the dialogue history for subsequent arbitration and generation. An \texttt{IGNORE} decision does not admit the triggering input as a new user turn. 
If the response has already been aborted by the fast interruption gate but the subsequent arbitration does not admit the input, the controller can use the saved assistant context to generate a continuation of the previous response.
For an \texttt{INTERRUPT\_AND\_REPLY} decision, the controller requests interruption if playback remains active, then admits the input as a new user turn and invokes the Reactive Dialogue Generator to produce a replacement response.

\subsection{Embodiment Cue and Routing Interface}
\label{subsec:motion_generation}
\label{subsec:tag_interface}
The embodiment cue $m$ is the explicit interface between CORTEX and physical behavior. It is serialized as a compact prefix:
\begin{equation}
<\texttt{motion:}\,m\,>;\text{response text},\qquad 
m\in\mathcal{M},
\end{equation}
where $\mathcal{M}$ is a finite vocabulary of embodiment cues, of which representative examples are shown in Tab.~\ref{tab:cue_vocab}. 
Embodiment-cue selection jointly considers the user's utterance, inferred affect, semantic intent, and recent dialogue history (as mentioned in Reactive Dialogue Generator). These signals are integrated to select a behavior appropriate to the conversational context, rather than applying a one-to-one mapping between affect and action. For example, negative affect may increase the likelihood of conversational repair and an \texttt{apologize} cue when the utterance indicates dissatisfaction, but does not independently determine the selected behavior. 

At execution time, a parser separates the cue prefix from the response text before passing the latter to TTS, preventing control metadata from entering spoken audio. If the cue is missing or malformed, the dispatcher falls back deterministically to \texttt{speak} for non-empty responses and \texttt{idle} for empty responses.

\begin{table}[t]
\centering
\small
\caption{\textbf{Representative embodiment-cue vocabulary.}
The table shows a representative subset of the cue vocabulary used in our system. Each cue specifies a communicative behavior family or execution route, which is realized through either streaming co-speech generation or pre-validated robot behaviors.}
\begin{tabularx}{\linewidth}{lXl}
\toprule
\textbf{Cue} & \textbf{Communicative role} & \textbf{Behavior family / route} \\
\midrule
\texttt{speak}
& General verbal response
& Streaming co-speech generation \\
\texttt{greeting}/\texttt{wave}
& Greeting or positive social opening
& Greeting / wave bank \\
\texttt{listen}
& Turn yielding or invitation to elaborate
& Attentive-listening bank \\
\texttt{confused}
& Uncertainty, misunderstanding, or failed understanding
& Confusion / clarification bank \\
\texttt{apologize}
& Conversational repair or explicit apology
& Apology / repair bank \\
\texttt{idle}
& Neutral hold or fallback
& Idle bank / default pose \\
\bottomrule
\end{tabularx}
\label{tab:cue_vocab}
\end{table}

\noindent\textbf{Embodiment-cue routing.}
The parsed embodiment cue then determines the downstream embodiment route. For non-speaking cues, CORTEX forwards the selected behavior family to a retrieval service containing pre-validated robot motions. These motions provide predictable execution timing and a bounded physical envelope for discrete social behaviors. For \texttt{speak}, CORTEX instead establishes a co-speech session and streams the associated TTS audio to ROSCO (Sec.~\ref{sec:generation}), which generates speech-synchronized motion online. This routing separates behavioral intent from motion realization: CORTEX determines what kind of embodied response is appropriate, while the corresponding execution module determines how that behavior is physically realized over time.

\subsection{Robot-Side Physical Execution}
\label{subsec:robot_execution}

\paragraph{Robot-side execution bridge and high-rate control.}
MIRA is deployed on an Astribot S1 humanoid robot. As a
robot-side component of MIRA, the deployed Astribot execution
bridge connects the system's behavior-generation modules to the
robot-side Orin service. It receives either discrete embodiment cues or streaming co-speech motion, with each motion stream represented as a sequence of qpos frames. The bridge translates these inputs into robot-side commands
and enforces the final physical execution constraints.

Discrete embodiment cues are mapped to pre-validated robot actions and dispatched to the robot.
For \texttt{speak}, the bridge opens a realtime co-speech session and receives a motion stream generated by ROSCO
(Sec.~\ref{sec:generation}). The first qpos frame of the stream is
sent to a brief \texttt{move\_to} transition.
This transition moves the robot smoothly from the pose produced by the preceding action, or from its current held pose, to the initial co-speech generated qpos frame, avoiding an abrupt discontinuity at the handoff.
Subsequent qpos frames are dispatched to the robot in realtime.


Before physical dispatch, each qpos frame in the generated motion
stream is passed through a collision-aware projection layer. The layer
evaluates the frame in the MuJoCo simulator~\footnote{https://mujoco.org/} and solves constrained inverse kinematics subject to joint-limit and collision constraints. If projection fails, the bridge holds the last safe pose or invokes the configured fallback policy. 
On the Orin side, source-rate motion frames, typically generated at 30~Hz, are consumed by a 250~Hz joint-position control loop. The control loop additionally enforces per-tick joint-step limits before forwarding commands to the Astribot SDK.

The execution bridge thus serves as the final execution and safety layer for physical behavior. When CORTEX interrupts a response, the active motion session is terminated and invalidated, causing any delayed commands from that session to be rejected. The controller then holds the last safe pose or invokes the configured fallback policy. This separation ensures that obsolete or unsafe motion cannot continue after an interruption, while leaving the dialogue state intact.

\subsubsection{Failure Containment and Recovery}
\label{subsec:failure_recovery}
As a streaming architecture composed of asynchronous modules and a physical execution loop, MIRA requires explicit failure isolation across module boundaries. Recovery is therefore handled locally, with each fault assigned a bounded scope determined by the responsible module. Language-format errors are contained before reaching TTS; interruptions are handled at the sentence level without invalidating the dialogue session; insufficient audio evidence prevents new motion from being committed; and collision rejection is handled independently of speech generation. Physical execution remains subject to the constraints enforced by the robot-side controller. The modular architecture also makes the individual service components replaceable. Streaming speech APIs, dialogue models, and downstream behavior services can be wrapped behind the session-scoped CORTEX interface without changing the cue-routing protocol.

\section{\rosco: Robotic Streaming Co-Speech Generation}
\label{sec:generation}

\begin{figure}[t]
  \centering
  \includegraphics[width=1.0\linewidth]{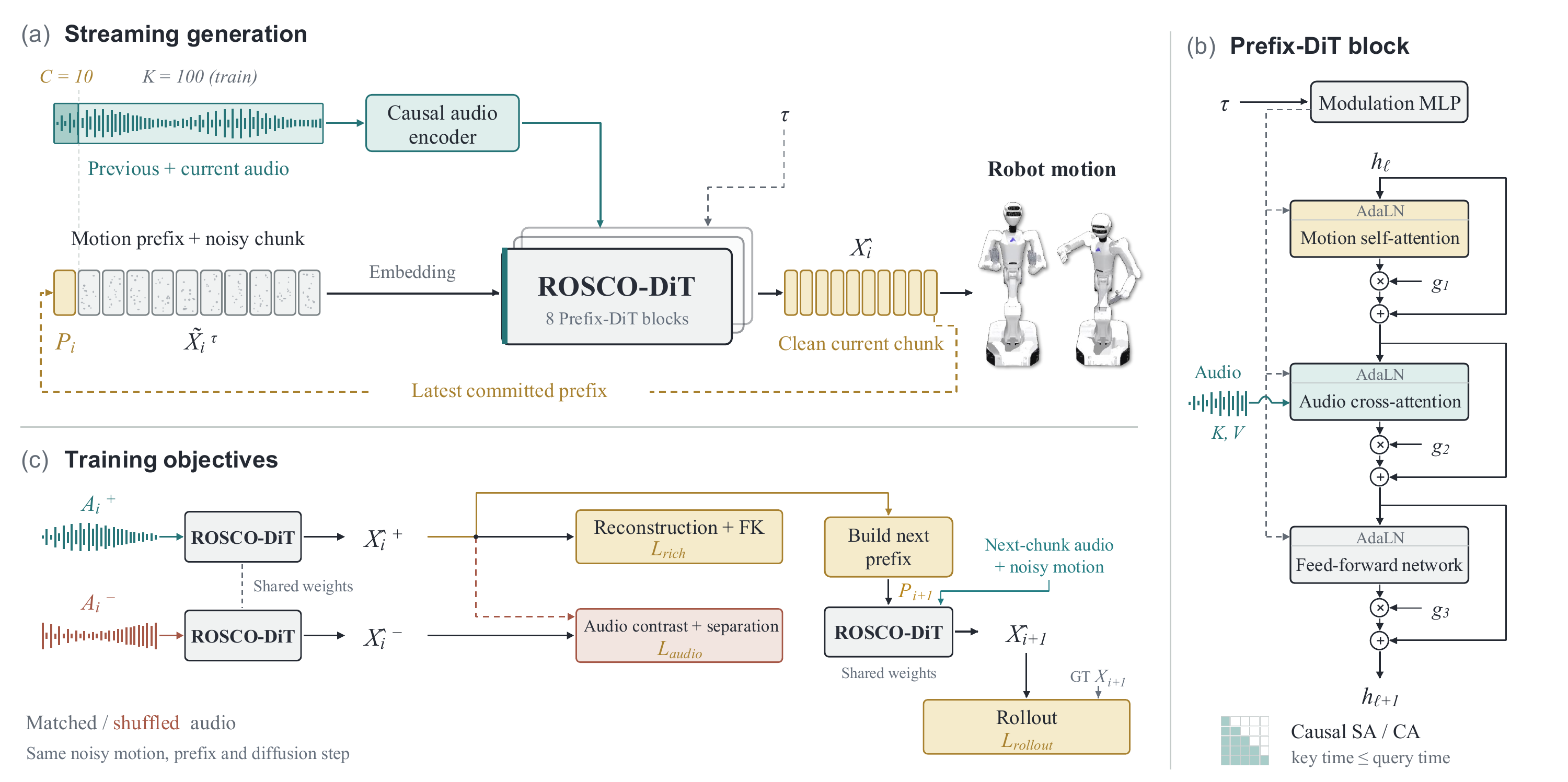}
  \caption{\textbf{Overview of ROSCO.}
(a) \textbf{Streaming generation.} ROSCO-DiT predicts a clean motion window conditioned on the preceding motion context and causal audio features. The cached rich-motion history provides the prefix for subsequent updates.
(b) \textbf{Prefix-DiT block.} ROSCO-DiT consists of stacked Prefix-DiT blocks, each applying causal motion self-attention, causal audio cross-attention, and a feed-forward network. A modulation MLP conditioned on the diffusion timestep $\tau$ generates the AdaLN shifts and scales, together with residual gates $g_1$, $g_2$, and $g_3$.
(c) \textbf{Training objectives.}
Matched and shuffled audio conditions share the same noisy motion, motion prefix, and diffusion timestep.
The matched prediction is supervised by the rich-motion reconstruction objective, including kinematic consistency and collision constraints, while a margin-based audio loss contrasts matched and shuffled conditions through reconstruction loss and joint-qpos prediction separation.
For autoregressive rollout, the matched prediction is detached, its joint velocities are recomputed from the predicted qpos, and the latest $C$ frames are retained to form the next prefix $P_{i+1}$.
The shared model then predicts the next chunk for rollout supervision.}
  \label{fig:prefix_dit}
\end{figure}

The \tagtoken{speak} route requires robot motion to remain synchronized with the temporal structure of speech, including its rhythm, pauses, and prosodic emphasis. Unlike retrieved behavior families, whose durations and trajectories are fixed in advance, co-speech motion must unfold continuously with the generated utterance, while future audio is not yet available. This makes online synchronization particularly challenging: motion must begin before the complete utterance is synthesized while remaining temporally consistent as new audio arrives. To address this challenge, we present \textbf{\rosco} (\textbf{Ro}botic \textbf{S}treaming \textbf{Co}-speech generator), a prefix-conditioned model that generates temporally coherent motion from streaming speech while preserving continuity across successive output chunks. The overview of ROSCO is presented in Fig.~\ref{fig:prefix_dit}.

\subsection{Motion Representation and Data Construction}
\label{subsec:motion_repr}

Training samples for \rosco are retargeted from co-speech dataset BEAT~\cite{liu2022beat} to Astribot S1 joint trajectories at 30~fps using GMR~\cite{araujo2025retargeting}. Each sample consists of a motion prefix of $C=10$ frames and a target motion chunk of $K=100$ frames, together with temporally aligned audio sequences.

To provide richer geometric supervision for motion reconstruction, we represent each motion frame using a rich motion representation comprising joint qpos, joint velocities, local-body positions, and local link rotations:
\begin{equation}
    x_t = [q_t,\; \Delta q_t,\; p_t^{\mathrm{local}},\; R_t^{6D}] \in \mathbb{R}^{460},
\end{equation}
where $q_t \in \mathbb{R}^{32}$ denotes the joint qpos sequence, $\Delta q_t \in \mathbb{R}^{32}$ the joint velocity, $p_t^{\mathrm{local}} \in \mathbb{R}^{132}$ the Cartesian positions of 44 tracked body segments, and $R_t^{6D} \in \mathbb{R}^{264}$ the local link rotations in the continuous 6D representation~\cite{zhou2019rotation}. Joint qpos and local-body positions are normalized using statistics computed over the training split, with the same statistics reused during inference. Joint velocities are computed from the unnormalized trajectories and scaled by the joint-position standard deviation. The auxiliary geometric components are used for training and reconstruction, while only the denormalized joint qpos $q_t$ are retained for downstream robot execution.

\subsection{ROSCO-DiT: Prefix-Conditioned Diffusion Transformer}
\label{subsec:rosco_dit}

ROSCO employs a prefix-conditioned Diffusion Transformer, termed \rosco-DiT, to generate each target motion chunk conditioned on preceding motion and temporally aligned audio. The \rosco-DiT backbone consists of eight stacked Prefix-DiT blocks, which extend the standard Diffusion Transformer architecture with explicit conditioning on preceding motion and its temporally aligned audio. 
Let $X_i \in \mathbb{R}^{K \times D_m}$ denote the target rich-motion chunk, $P_i \in \mathbb{R}^{C \times D_m}$ the preceding motion prefix, $A_i \in \mathbb{R}^{K \times D_a}$ the audio features aligned with the target chunk, and $A_i^{\mathrm{prev}} \in \mathbb{R}^{C \times D_a}$ the audio features aligned with the prefix. Following the standard diffusion formulation, Gaussian noise is added to the target motion $X_i$ at diffusion timestep $\tau$:
\begin{equation}
    \tilde{X}_i^{\tau} = \sqrt{\bar{\alpha}_{\tau}} X_i +
    \sqrt{1-\bar{\alpha}_{\tau}}\epsilon,\quad
    \epsilon\sim\mathcal{N}(0,I).
\end{equation}
The model then reconstructs the clean target motion conditioned on the motion prefix and aligned audio:
\begin{equation}
    \hat{X}_i =
    f_{\theta}(\tilde{X}_i^{\tau}, P_i, A_i, A_i^{\mathrm{prev}}, \tau),
\end{equation}
where $\tau$ is the diffusion timestep.
The noisy target tokens $\tilde{X}_i^{\tau}$ and clean prefix tokens $P_i$ are concatenated along the temporal dimension and marked with a binary type indicator distinguishing prefix frames from target frames. The resulting motion sequence is projected to $d=256$ dimensions and augmented with sinusoidal temporal position embeddings. To support streaming inference, temporal information flow is constrained to be causal at both the audio encoding and motion generation stages. Specifically, the prefix and target audio sequences are processed by a shared causal audio encoder, producing $E_i^{\mathrm{prev}}$ and $E_i$. Each transformer block then applies causally masked self-attention over motion tokens, followed by causal audio cross-attention to the encoded audio, where each motion token can access only audio tokens at the same or earlier temporal positions, and a feed-forward network. The model uses eight transformer blocks with four attention heads, a 1024-dimensional feed-forward layer, and dropout of 0.1. The output head removes the prefix portion and predicts only the denoised target chunk $\hat{X}_i$.

The diffusion timestep $\tau$ is embedded and injected into each block through adaptive LayerNorm (AdaLN). Separate modulation parameters are used for the self-attention, audio cross-attention, and feed-forward branches, including shift, scale, and residual gates.

\subsection{Joint Training Objectives and Contrastive Loss}
\label{subsec:training_objectives}

The training objective combines rich-motion reconstruction, kinematic consistency, autoregressive rollout, and acoustic conditioning. For clarity, we omit the batch and chunk index $i$ in the following formulation.

The model is trained with a global reconstruction loss over the rich motion representation, together with component-wise losses that explicitly emphasize the joint configuration, velocity, local-body position, and rotation components, as well as additional kinematic constraints:
\begin{equation}
\begin{split}
    \mathcal{L}_{\mathrm{rich}} =
    &\lambda_x\operatorname{MSE}(\hat X,X)
    +\lambda_q\operatorname{MSE}(\hat q,q)
    +\lambda_{\Delta q}\operatorname{MSE}(\widehat{\Delta q},\Delta q) \\
    &+\lambda_p\operatorname{MSE}(\hat p^{\mathrm{local}},p^{\mathrm{local}})
    +\lambda_R\operatorname{MSE}(\hat R^{6D},R^{6D})
    +\lambda_{\mathrm{kin}}\mathcal{L}_{\mathrm{kin}},
\end{split}
\end{equation}
where $\lambda_x$ weights the reconstruction of the complete rich-motion representation, while $\lambda_q$, $\lambda_{\Delta q}$, $\lambda_p$,  $\lambda_R$, and $\lambda_{\mathrm{kin}}$ control the contributions of joint qpos, joint velocities, local-body positions, local link rotations, and kinematic constraints, respectively. $\hat \cdot$ denotes the predicted component.
To constrain the predicted joint qpos, we incorporate robot-specific kinematic and collision constraints by mapping the predictions to Cartesian positions through forward kinematics (FK) and explicitly penalizing close-proximity self-collisions.
Let $\operatorname{FK}(\hat q_t)$ denote the forward kinematics function that maps the predicted joint qpos to the Cartesian positions of the tracked body segments, and let $p_t^{\mathrm{FK}}=\operatorname{FK}(\hat q_t)$ denote its output. We define:
\begin{equation}
    \mathcal{L}_{\mathrm{kin}}=
    \lambda_{\mathrm{FK\text{-}pos}}\operatorname{MSE}(p^{\mathrm{FK}},p^{\mathrm{local}})
    +\lambda_{\mathrm{col}}\frac{1}{T}\sum_t
    \left[m_{\mathrm{col}}-\operatorname{Clearance}(\hat q_t)\right]_+^2,
\end{equation}
where $\operatorname{Clearance}(\hat q_t)$ denotes the minimum pairwise clearance between the collision proxies of the body segments induced by the predicted qpos. The first term enforces Cartesian consistency through forward kinematics, while the second encourages a minimum clearance and penalizes potential self-collisions.

Moreover, to mitigate exposure bias during recurrent inference, we additionally perform a two-step autoregressive rollout during training. The first predicted chunk is detached and used as the motion prefix for a subsequent target chunk $X'$, with joint velocities recomputed from the predicted qpos. The rollout loss uses the same rich-motion objective:
\begin{equation}
    \mathcal{L}_{\mathrm{rollout}}
    =\mathcal{L}_{\mathrm{rich}}(\hat X',X'),
\end{equation}
where $\hat X'$ denotes the prediction conditioned on the generated motion prefix.

To prevent the model from over-relying on motion history and ignoring acoustic cues, we construct negative conditions by cyclically shuffling the current and prefix audio across each training batch. Let $\mathcal{L}^{+}_{rich}$ and $\mathcal{L}^{-}_{rich}$ denote the rich-motion reconstruction losses under matched and shuffled audio conditions, respectively. We encourage the model to assign a higher reconstruction loss to mismatched audio while requiring the corresponding motion predictions to respond differently:
\begin{equation}
\begin{split}
    \mathcal{L}_{\mathrm{audio}} =
    &\left[
    \operatorname{sg}(\mathcal{L}^{+}_{rich})
    +m_{\mathrm{audio}}
    -\mathcal{L}^{-}_{rich}
    \right]_+ \\
    &+\lambda_{\mathrm{sep}}
    \left[
    m_{\mathrm{sep}}
    -\operatorname{mean}|\hat q^{+}-\hat q^{-}|
    \right]_+,
\end{split}
\end{equation}
where $\hat q^{+}$ and $\hat q^{-}$ denote the predicted joint qpos under matched and shuffled audio conditions, respectively. 
The margin $m_{\mathrm{audio}}$ encourages a margin between the reconstruction losses under matched and mismatched audio, with matched audio expected to yield a lower loss. In contrast, $m_{\mathrm{sep}}$ encourages a minimum difference between the corresponding motion predictions in joint qpos space.
The stop-gradient operator $\operatorname{sg}(\cdot)$ prevents the matched-condition loss from being optimized through this comparison.
The complete training objective is:
\begin{equation}
    \mathcal{L}=
    \lambda_{\mathrm{rich}}\mathcal{L}_{\mathrm{rich}}
    +\lambda_{\mathrm{roll}}\mathcal{L}_{\mathrm{rollout}}
    +\lambda_{\mathrm{audio}}\mathcal{L}_{\mathrm{audio}}.
\end{equation}

\subsection{Receding-Horizon Prefix Commitment Inference}
\label{subsec:inference_tricks}

\begin{figure}[t]
\centering
\includegraphics[width=1.0\linewidth]{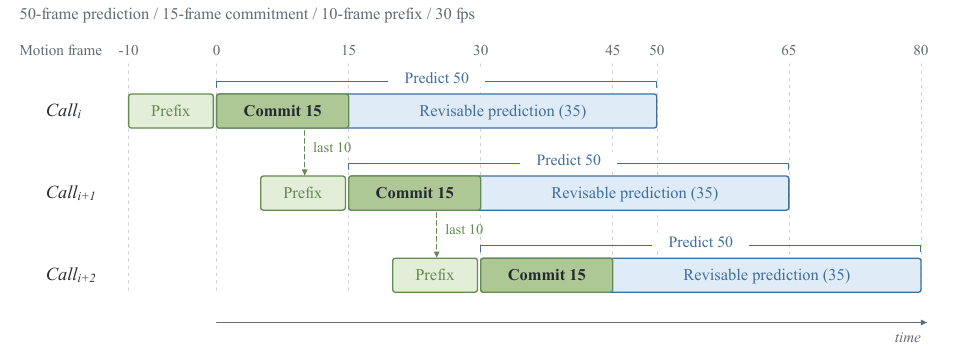}
\caption{\textbf{Overlapping-window streaming inference.} Each call predicts 50 frames but commits only the first 15. The next call advances by 15 frames, reuses the latest 10 committed frames as prefix context, and predicts an overlapping future horizon.}
\label{fig:streaming_window}
\end{figure}

To balance the need for a sufficiently long temporal context with the low-latency requirements of streaming robot execution, we design \textbf{Receding-Horizon Prefix Commitment (RHPC)}, as shown in Fig.~\ref{fig:streaming_window}. Rather than generating and committing short motion chunks independently, RHPC maintains a longer prediction horizon while committing only a short leading segment at each inference step.
Since no ground-truth motion history is available at test time, inference in \rosco differs from offline clip prediction. Specifically, at the start of a speaking turn, the streaming bridge initializes the motion prefix $P_0 \in \mathbb{R}^{C \times D_m}$ and corresponding audio context $A_0^{\mathrm{prev}}$ with zero tensors. The generated motion chunk is then fed back as the prefix for the subsequent inference step, enabling autoregressive streaming generation.

At each inference step, the model consumes 50 temporally aligned audio tokens and predicts an extended 50-frame rich-motion window, but only the first 15 frames are committed to the robot controller. At the next inference step, the model predicts the next 50-frame window using the latest 10 committed frames as the motion prefix, with the commitment progressing by 15 frames at each step. The 10-frame overlap improves continuity between newly generated and previously committed motion, resulting in smoother transitions across chunk boundaries. 
Maintaining this 50-frame prediction horizon serves two practical purposes. First, it preserves a temporal sequence length closer to that used during training, reducing the train-test distribution mismatch that would arise from applying diffusion denoising to substantially shorter sequences. Second, it provides an extended trajectory candidate that allows downstream kinematic projection and collision checks to assess future motion feasibility before commands are dispatched.

The 15-frame commitment also bounds the amount of motion exposed to the robot at each inference step, which is beneficial for full-duplex interaction. Since only 15 frames (0.50~s) are committed at a time, an interruption triggered by CORTEX does not require discarding a long pre-generated trajectory. Instead, the active motion session can be terminated after the currently committed segment, while continuity is preserved up to the last committed frame and the system remains responsive to user barge-in.



\section{Experiments}
\label{sec:eval}

We evaluate MIRA as an end-to-end embodied interaction system along three complementary dimensions: (i) co-speech motion quality, (ii) real-time streaming performance, (iii) contextual turn arbitration on real interaction traces. The evaluation combines component-level measurements with end-to-end experiments on a physical humanoid platform.

\begin{figure}
    \centering
    \includegraphics[width=\linewidth]{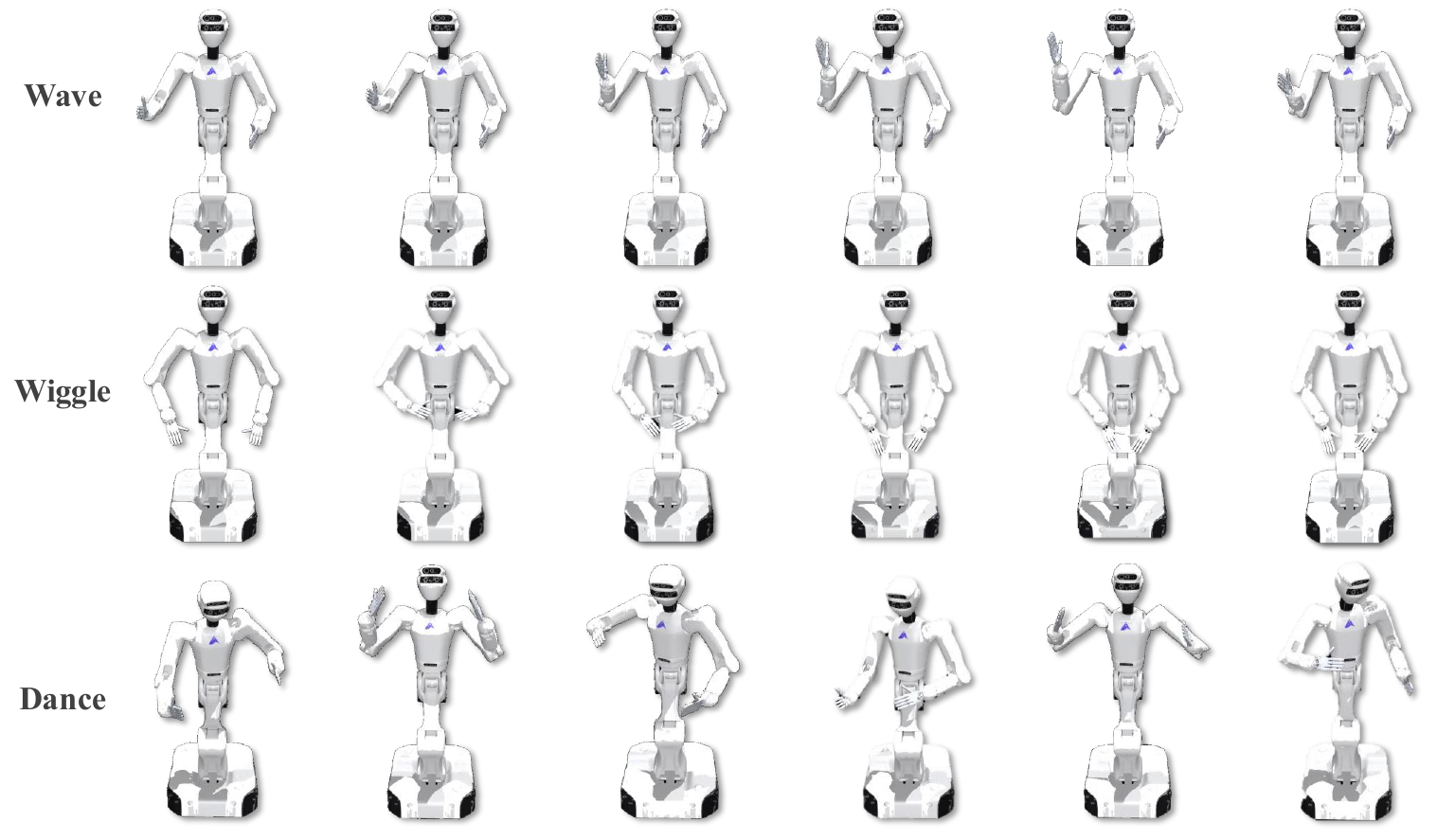}
    \caption{\textbf{Visualization of representative robot motions in the embodiment-cue vocabulary.} Each embodiment cue is associated with multiple pre-authored motion chunks, from which one is randomly selected during interaction.}
    \label{fig:tag_visualize}
\end{figure}

We conduct all physical experiments on an Astribot~S1 deployment integrating CORTEX, the validated behavior service, the \rosco co-speech bridge, and the robot-side controller. Unless otherwise specified, all experiments use the same trained model and deployment configuration. We visualize several representative robot motions from the embodiment-cue vocabulary in Fig.~\ref{fig:tag_visualize}.

\subsection{Implementation Details}
\label{subsec:implementation_details}

\rosco operates at 30~fps with 16~kHz mono audio. Each training sample contains a $C=10$-frame motion prefix and a $K=100$-frame target chunk, corresponding to approximately $0.33$~s and $3.33$~s, respectively. Samples near the beginning of a clip are left-padded with the initial motion frame to obtain a fixed-length prefix. The rich-motion representation has dimension $D_m=460$, comprising 32 joint qpos, 32 joint velocities, 132 local-body position values, and 264 local link rotations represented using the continuous 6D representation~\cite{zhou2019rotation}. Only the denormalized joint qpos are used for downstream robot execution.

For each motion frame, the audio descriptor consists of 80 log-mel channels and four low-latency prosodic features---log energy, delta log energy, positive onset strength, and zero-crossing rate---giving $D_a=84$. The audio features are processed by a causal dilated encoder with dilation rates $[1,2,4,8,16]$, shared between the prefix and target audio streams. The ROSCO-DiT projects motion tokens to $d=256$ dimensions and uses eight Prefix-DiT blocks with four attention heads and a 1024-dimensional feed-forward layer. Sinusoidal temporal position embeddings are applied to the motion tokens, while the diffusion timestep is injected into each transformer block through AdaLN. The diffusion process follows the standard DDPM noise schedule~\cite{ho2020ddpm}.

To improve robustness to the incomplete and self-generated context encountered during streaming inference, we apply condition corruption during training and perform a two-step autoregressive rollout. Audio conditions are independently masked, while the motion prefix is randomly removed, truncated, or perturbed with Gaussian noise. In the autoregressive rollout, the motion generated in the first prediction is detached and reused as the prefix for the second prediction, with joint velocities recomputed from the generated joint qpos. The second-step rollout loss is weighted by $\lambda_{\mathrm{roll}}=0.5$. These procedures expose the model to the imperfect conditioning states that arise during streaming deployment.

At deployment, \rosco performs causal inference using a 50-frame temporal window with a 15-frame commitment. Incoming TTS audio is resampled from 24~kHz to 16~kHz mono before entering the inference stream. The inference procedure follows the Receding-Horizon Prefix Commitment strategy described in Sec.~\ref{subsec:inference_tricks}. Only the denormalized joint qpos of the committed 15 frames are transmitted to the robot controller, while the remaining rich-motion predictions are retained internally for subsequent inference.

\subsection{\rosco Motion Quality Evaluation}
\label{subsec:generation_eval}

\noindent\textbf{Dataset and Comparison Methods.}
We evaluate the algorithmic motion quality of \rosco on the retargeted BEAT~\cite{liu2022beat} test set. We compare against representative co-speech motion generation methods, including GestureLSM~\cite{liu2025gesturelsm} and MIBURI~\cite{mughal2026miburi}. For all comparison methods, the generated motions are retargeted to the Astribot S1 using the same retargeting protocol as described in Sec.~\ref{subsec:motion_repr}, \ie GMR~\cite{araujo2025retargeting}. We evaluate three complementary aspects of generated motion: audio-motion synchronization, motion-distribution fidelity, and physical executability. All metrics are computed from robot kinematic states (\ie FK positions) obtained by applying \texttt{mj\_forward} to the predicted or retargeted qpos trajectories.\\
\noindent\textbf{Audio-Motion Synchronization.}
For audio-motion synchronization, we report two complementary beat-level synchronization metrics: Beat Consistency (BC)~\cite{liu2025gesturelsm} and BeatAlign~\cite{huang2026omg}. Since the two metrics use different beat extraction and alignment procedures, we report them separately rather than treating their scores as directly interchangeable.
For BC, let $A=\{a_i\}$ denote the audio-onset timestamps detected from the input waveform, where each $a_i$ marks the temporal location of a detected audio onset. Let $p_{t,j}\in\mathbb{R}^{3}$ denote the FK position of the $j$-th selected joint at frame $t$. We compute the frame-to-frame speed of each selected joint and detect local minima using an order of seven frames. We then retain local minima with normalized speed above $0.3$, producing an independent motion-beat set $M_j$ for each joint. For each audio onset $a\in A$, we find its nearest motion beat in $M_j$ and apply a Gaussian temporal weighting:
\begin{equation}
    \mathrm{BC}_j
    =
    \frac{1}{|A|}
    \sum_{a\in A}
    \exp\left(
        -\frac{d(a,M_j)^2}{2\sigma_{\mathrm{E}}^2}
    \right),
    \qquad
    d(a,M_j)=\min_{m\in M_j}|a-m|,
\end{equation}
where $\sigma_{\mathrm{E}}=0.3\,\mathrm{s}$. The final BC score is obtained by averaging the per-joint scores over the $J$ selected joints:
\begin{equation}
    \mathrm{BC}
    =
    \frac{1}{J}\sum_{j=1}^{J}\mathrm{BC}_j.
\end{equation}
Thus, BC measures audio-to-motion temporal alignment independently for each selected joint, without merging motion beats across joints.
For BeatAlign, we instead construct a single global motion signal by averaging the frame-difference speed over the $J$ selected joints:
\begin{equation}
    u_t
    =
    \frac{1}{J}
    \sum_{j=1}^{J}
    \|p_{t+1,j}-p_{t,j}\|_2.
\end{equation}
We detect local minima of $u_t$, requiring adjacent motion beats to be at least $0.25\,\mathrm{s}$ apart, yielding a single global motion-beat set $M$. BeatAlign then computes the audio-to-motion alignment as
\begin{equation}
    \mathrm{BeatAlign}(A,M)
    =
    \frac{1}{|A|}
    \sum_{a\in A}
    \exp\left(
        -\frac{d(a,M)^2}{2\sigma_{\mathrm{B}}^2}
    \right),
    \qquad
    d(a,M)=\min_{m\in M}|a-m|,
\end{equation}
where $\sigma_{\mathrm{B}}=3$ motion frames $=0.1\,\mathrm{s}$ at $\mathrm{FPS}=30$. This score averages over audio onsets and therefore measures how well the generated robot motion covers the detected audio onsets.\\
\noindent\textbf{Motion-Distribution Fidelity.}
To measure motion-distribution fidelity, we compute FID-G distance. For each prediction-reference pair, we first truncate both
trajectories to their common duration and resample them to a common frame
rate. We remove global translation by expressing each selected body
position relative to the root body. For each clip, we summarize each
root-local position dimension over time using its mean, standard deviation,
5-th percentile, and 95-th percentile, yielding a fixed-dimensional
geometry feature vector. We then compute the Fr\'echet distance between the
predicted and reference feature distributions:
\begin{equation}
    \mathrm{FID\text{-}G}
    =
    \|\mu_p-\mu_r\|_2^2
    +
    \operatorname{Tr}\left(
        \Sigma_p+\Sigma_r
        -2(\Sigma_p\Sigma_r)^{1/2}
    \right),
\end{equation}
where $\mu_p$ and $\Sigma_p$ denote the empirical mean vector and
covariance matrix of the predicted geometry features, respectively, and
$\mu_r$ and $\Sigma_r$ denote those of the reference features. Lower FID-G indicates closer alignment with the reference motion distribution.\\
\noindent\textbf{Kinematic self-collision proxy.}
To assess simulation-based feasibility, we replay every predicted qpos frame in the full Astribot MuJoCo model and call \texttt{mj\_forward} without stepping the dynamics. Thus, the supplied trajectory is evaluated kinematically, without modification by gravity, contacts, or actuators. We count contacts with non-positive signed distance and distinguish self-collisions from environment collisions using MuJoCo body identifiers. We report the self-collision frame rate, defined as the fraction of frames containing at least one self-contact. This simulator-based measure serves as a proxy for physical executability, but does not by itself guarantee successful hardware execution.

\begin{table}[t]
\centering
\small
\caption{\textbf{Quantitative evaluation of \rosco.} BC and BeatAlign measure audio-motion synchronization, FID-G measures geometric motion-distribution discrepancy, and Collision Rate measures the rate of kinematic self-collision. Bold values indicate the best performance among generative baselines.}
\begin{tabular*}{\linewidth}{l@{\extracolsep{\fill}}cccc}
\toprule
Method & 
\shortstack{BC $\uparrow$} & 
\shortstack{BeatAlign $\uparrow$} & 
\shortstack{FID-G $\downarrow$} & 
\shortstack{Collision Rate (\%) $\downarrow$} \\
\midrule
Ground Truth (GT)
& $0.2536$ & $0.5600$ & -- & $6.51\%$ \\

GestureLSM~\cite{liu2025gesturelsm}
& $0.0049$ & $\mathbf{0.5849}$ & $0.3420$ & $3.09\%$ \\

MIBURI~\cite{mughal2026miburi}
& $0.0766$ & $0.5552$ & $0.2427$ & $3.01\%$ \\
\textbf{\rosco (Ours)}
& $\mathbf{0.3039}$ & $0.5726$ & $\mathbf{0.1111}$ & $\mathbf{2.61\%}$ \\
\bottomrule
\end{tabular*}
\label{tab:rhythmic_alignment}
\end{table}

\noindent\textbf{Evaluation Results.}
As shown in Table~\ref{tab:rhythmic_alignment}, \rosco achieves the highest BC score among the generative baselines, indicating stronger rhythmic synchronization between the detected audio onsets and generated motion beats. It also remains competitive on BeatAlign, where a higher score indicates better temporal alignment between motion dynamics and the rhythmic structure of the input audio. This suggests that \rosco can effectively capture audio-driven motion timing beyond simple frame-wise correspondence. In addition, \rosco obtains the lowest FID-G among the reported generative models, indicating that its generated motion better matches the geometric distribution of real motion.

Notably, \rosco also achieves a low collision rate, substantially lower than the retargeted ground-truth motions. This difference is expected because the GMR retargeting process primarily aims to preserve the kinematic correspondence between the source and target embodiments, without explicitly optimizing for collision avoidance. As a result, the retargeted reference motions can still contain physically undesirable configurations, as reflected by the relatively high collision rate of 6.51\% for GT. In contrast, \rosco incorporates a collision-aware training objective that explicitly penalizes physically implausible motions during generation. This encourages the model to learn motions that balance audio-motion alignment with the physical constraints of the target robot, rather than simply reproducing the potentially colliding retargeted trajectories. Together with its strong BeatAlign and BC scores, the results suggest that \rosco can preserve audio-driven rhythmic and temporal correspondence while producing motion that is more compatible with robot execution.

\begin{table}[t]
\centering
\small
\caption{\textbf{Runtime latency and full-duplex responsiveness of MIRA on Astribot S1.} Evaluated across multi-turn interaction sessions. \texttt{user\_turn} denotes the committed user transcript post-ASR. All metrics report the median (P50).}
\label{tab:runtime_core}
\begin{tabularx}{\linewidth}{>{\raggedright\arraybackslash}X l c}
\toprule
\textbf{Pipeline Stage / Metric} & \textbf{Measurement Window} & \textbf{Median (P50)} \\
\midrule
\multicolumn{3}{l}{\textit{\textbf{Upstream Cognitive Breakdown}}} \\
\quad Streaming ASR Endpointing & \texttt{speech\_end} $\to$ \texttt{user\_turn} & 10.0 ms \\
\quad LLM First-Token Latency (TTFT) & \texttt{user\_turn} $\to$ \texttt{llm\_first\_token} & 1.62 s \\
\quad Streaming Audio Ingress & \texttt{llm\_first\_token} $\to$ \texttt{pcm\_received} & 493 ms \\
\midrule

\multicolumn{3}{l}{\textit{\textbf{Streaming Motion Generation}}} \\
\quad ROSCO Chunk Inference Time ($T_{\text{gen}}$) & \texttt{pcm\_received} $\to$ \texttt{motion\_ready} & 195 ms \\
\quad Emission Real-Time Factor (RTF) & $T_{\text{gen}} / T_{\text{motion}}$ ($T_{\text{motion}} = 500\text{ ms}$) & 0.390 \\
\midrule

\multicolumn{3}{l}{\textit{\textbf{Interactive Responsiveness}}} \\
\quad First Synthesized Speech & \texttt{user\_turn} $\to$ \texttt{tts\_first\_audio} & 2.20 s \\
\quad Co-Speech Motion Synchronization & \texttt{user\_turn} $\to$ \texttt{first\_motion\_udp} & 2.36 s \\
\midrule

\multicolumn{3}{l}{\textit{\textbf{Full-Duplex Preemption}}} \\
\quad Deterministic Interruption Preemption & \texttt{speech\_started} $\to$ \texttt{abort\_cmd\_sent} & 466~ms \\
\bottomrule
\end{tabularx}
\end{table}

\subsection{Real-Time Generation and Full-Duplex Responsiveness}
\label{sec:experiments_runtime}

Continuous human-robot interaction imposes three temporal requirements on the embodied system: \textbf{(i) Streaming Efficiency}: motion synthesis must run faster than physical playback to support continuous motion execution; \textbf{(ii) Turn Responsiveness}: the system should minimize the delay before speech and motion begin; and \textbf{(iii) Full-Duplex Preemption}: ongoing speech and physical execution must be promptly interruptible upon user barge-in. We benchmark these three aspects on the physical Astribot S1 platform across continuous multi-turn interaction sessions. Table~\ref{tab:runtime_core} summarizes the median (P50) runtime measurements.

\textbf{Streaming Motion Generation Efficiency.}
Streaming viability requires that the time to generate a motion chunk, denoted by $T_{\text{gen}}$, be shorter than the physical duration of the committed motion, denoted by $T_{\text{motion}}$, i.e., $T_{\text{gen}} < T_{\text{motion}}$ or equivalently $\text{RTF} < 1.0$. With a $T_{\text{motion}} = 500\text{ ms}$ committed motion duration, ROSCO completes prefix-conditioned diffusion inference in only 195~ms (\texttt{pcm\_received} $\to$ \texttt{motion\_ready}), yielding an emission RTF of 0.390. This indicates that motion generation operates substantially faster than physical playback, leaving sufficient runtime margin to sustain continuous trajectory execution under the evaluated deployment conditions.

\textbf{Turn Startup and Audio–Motion Coupling.}
In conversational turn transitions, the component breakdown shows that startup latency is dominated by upstream cognitive processing rather than motion synthesis. Local ASR endpointing introduces only 10.0~ms (\texttt{speech\_end} $\to$ \texttt{user\_turn}), while LLM first-token generation (1.62~s) and initial audio streaming to PCM ingress (493~ms) contribute most of the latency before first speech emission (\texttt{tts\_first\_audio} at 2.20~s). Once audio becomes available, ROSCO generates the initial joint trajectory within 195~ms of audio ingress, with motion dispatched to the hardware bridge at 2.36~s (\texttt{first\_motion\_udp}). 

\textbf{Full-Duplex Interruption Preemption.}
MIRA combines RHPC with a dedicated VAD-based interruption gate (\ie Fast Interruption Gate) to support prompt barge-in handling. RHPC commits only the leading 15 frames (500~ms) of each 50-frame prediction horizon, thereby strictly bounding the amount of motion committed ahead of physical execution. When user vocalization is detected during robot speech, the interruption gate requests active-audio cancellation and dispatches an asynchronous abort command. As reported in Table~\ref{tab:runtime_core}, the preemption latency is \textbf{466~ms} at the median, measured from speech onset (\texttt{speech\_started}) to abort-command dispatch (\texttt{abort\_cmd\_sent}). This bounded commitment mechanism limits the amount of precomputed motion that can remain committed after a user barge-in, while enabling subsequent turn arbitration to proceed without waiting for the ongoing motion sequence to complete.

\subsection{Multi-Party Turn Arbitration and Addressivity}
\label{subsec:turn_arbitration}
We evaluate CORTEX in unconstrained multi-party interaction using a production trace corpus containing 199 active sessions, 36,448 recorded events, and 559 completed turn decisions. Among these decisions, 302 resulted in \texttt{REPLY}, 88 in \texttt{INTERRUPT\_AND\_REPLY}, and 169 in \texttt{IGNORE}. Rather than evaluating isolated interruption examples, this trace-based evaluation examines whether CORTEX can distinguish speech addressed to the robot from concurrent third-party speech and decide whether an ongoing response should be continued or interrupted. Four representative cases are summarized in Table~\ref{tab:case_studies}.

\begin{table}[t]
\centering
\caption{\textbf{Representative multi-party turn-arbitration cases.} The examples show how the arbiter uses dialogue context for unclear cases and predefined rules for clear cases. The Fast Interruption Gate provides a separate fast path for physical interruption.}
\begin{tabularx}{\linewidth}{ccp{3.8cm}l}
\toprule
Case Family & User Utterance & Decision Basis & Decision \\
\midrule

\textbf{Case A}
& ``Just sent the location to them.''
& Dialogue context
& \texttt{IGNORE} \\

\addlinespace

\textbf{Case B}
& ``No, that will not work.''
& Dialogue context
& \texttt{INTERRUPT\_AND\_REPLY} \\

\addlinespace

\textbf{Case C}
& ``Um.''
& Predefined rule
& \texttt{IGNORE} \\

\addlinespace

\textbf{Case D}
& ``MIRA.''
& Predefined rule
& \texttt{INTERRUPT\_AND\_REPLY} \\

\bottomrule
\end{tabularx}
\label{tab:case_studies}
\end{table}

\textbf{Non-addressed speech.} Case A demonstrates CORTEX's ability to handle multi-party overhearing and recover gracefully from false-alarm early aborts. While the robot is presenting, a bystander says ``Just sent the location to them'' to another person, with vocalization exceeding the 450~ms threshold. Because physical preemption prioritizes low-latency halting, the Fast Interruption Gate triggers an \texttt{early\_barge\_in\_abort}, temporarily pausing active speech and holding the robot motion. Once the finalized ASR transcript arrives, the Deliberative Turn Arbiter evaluates the preceding conversational context and determines the utterance was not addressed to the robot, returning \texttt{IGNORE}. Rather than discarding interaction progress or terminating awkwardly, CORTEX retrieves the saved response context and seamlessly generates a continuation from the interrupted breakpoint, successfully recovering from the environmental false-alarm.

\noindent\textbf{Context-dependent interruption.}
Case~B demonstrates how CORTEX handles a short utterance whose intent is unclear from the utterance alone. The user says ``No, that will not work'' while the robot is presenting a proposal. The utterance itself does not provide enough information to determine the appropriate response, so the Deliberative Turn Arbiter uses the preceding dialogue context to infer the user's intent and returns \texttt{INTERRUPT\_AND\_REPLY}. Importantly, the physical preemption does not wait for this decision. In this trace, the Fast Interruption Gate emitted an \texttt{early\_barge\_in\_abort} event before the finalized ASR transcript reached $\mathcal{P}_{\mathrm{arb}}$. This separation allows the robot to stop promptly while the Deliberative Turn Arbiter continues to determine how to respond to the interruption.

\textbf{Rule-based decisions.} Cases~C and~D show that clear inputs can be handled directly by predefined rules in the Deliberative Turn Arbiter. In Case C, the user produces a brief hesitation filler (``Um'', duration $< 450$~ms) during playback. Because the utterance does not satisfy the sustained 450~ms confirmation threshold, the Fast Interruption Gate is not engaged. Upon ASR finalization, the arbiter matches the filler rule and returns \texttt{IGNORE} within $\sim$10~ms, allowing continuous robot playback to proceed completely undisturbed. In Case D, the user issues an explicit wake word (``MIRA''). The Fast Interruption Gate initiates the hardware abort at 450~ms, and the arbiter confirms the addressivity via wake-word detection, returning \texttt{INTERRUPT\_AND\_REPLY} only $2$~ms after the gate's decision.

Together, these cases demonstrate how CORTEX combines rule-based decisions with context-based reasoning within the Deliberative Turn Arbiter. Clear inputs, such as fillers and explicit wake-word commands, can be handled directly by predefined rules, while short or ambiguous utterances can be interpreted using the preceding dialogue context. The Fast Interruption Gate handles the time-critical physical stop separately, allowing the robot to stop promptly without waiting for the dialogue-level decision. This design enables CORTEX to avoid unnecessary interruptions while responding quickly when an interruption is clearly required.

\section{Discussion and Limitations}
\label{sec:discussion}

\textbf{System Insights: Hybrid Embodiment and Streaming Quality.}
MIRA couples dialogue reasoning, motion generation, and physical execution into a unified interactive loop. Rather than treating motion as a passive downstream rendering step, our hybrid routing separates discrete social actions from open-ended speaking: pre-validated motion libraries provide deterministic safety envelopes for repeatable behaviors (e.g., greetings, listening), while streaming diffusion dynamically adapts to unpredictable speech prosody. In continuous physical deployment, streaming quality emerges as an end-to-end system property. Beyond the generative capacity of the diffusion backbone, real-time fluidity depends critically on prefix feedback, receding-horizon commitment schedules, boundary preemption, and robot-side execution safeguards.

\noindent\textbf{Limitations and Future Scope.} The current framework has several practical limitations that outline avenues for future investigation: 
(1) the embodiment vocabulary is currently constrained, where incorporating joint semantic text conditioning alongside audio prosody could further enrich gesture expressivity; 
(2) the physical safety layer requires embodiment-specific kinematic tuning and collision geometry modeling for new robot platforms; 
and (3) turn startup latency remains largely bounded by the upstream cascaded pipeline (ASR $\rightarrow$ LLM $\rightarrow$ TTS), where integrating native end-to-end speech-to-speech (omni) foundation models represents a promising avenue to substantially compress initial response delay. 
Finally, while our quantitative benchmarks establish the essential technical viability, streaming throughput, and physical containment required for embodied interaction, long-term human-subject studies in naturalistic environments remain an important next step to fully assess subjective companion dynamics and sustained user engagement.

\section{Conclusion}
\label{sec:conclusion}

We presented \systemname, a unified framework for real-time embodied companion interaction. MIRA links user-intent understanding, response generation, embodiment-cue prediction, and robot motion within a single streaming architecture. CORTEX maintains the interaction state, starts streaming responses, manages deliberative turn decisions, and handles bounded-latency interruption. Validated behavior families provide discrete social actions, while a prefix-conditioned diffusion model ROSCO generates open-ended co-speech motion from incremental audio and generated motion history.
During streaming inference, RHPC maintains a long prediction horizon for smooth streaming motion while limiting physical commitment to a short leading segment. Finally, a robot-side bridge provides a final safety layer, enforcing timing and physical constraints for safe execution. Together, these components form an extensible architecture that translates companion responses and embodiment cues into timely, expressive, and physically contained robot behavior.

\bibliographystyle{ACM-Reference-Format}
\bibliography{main}

\end{document}